\documentclass[letterpaper, 10 pt, conference]{ieeeconf}   
\usepackage{graphicx}
\usepackage{booktabs} 
\usepackage{braket}
\usepackage{hyperref}
\usepackage{multirow}
\usepackage{xcolor}
\usepackage{float}
\usepackage{cuted}
\usepackage[hypcap=true]{caption}
\usepackage{cuted}
\usepackage{stfloats}
\usepackage{lipsum}
\usepackage{subcaption}
\usepackage{amsmath}
\usepackage{colortbl}
\usepackage{multirow}
\usepackage{makecell}
\usepackage{hhline} 
\usepackage{algorithm}
\usepackage{algorithmic}
\usepackage{amsmath, amssymb}

\newcommand{\RGBinput}{\mathbf{R} \in \mathbb{R}^{H \times W \times 3}}
\newcommand{\Depthinput}{\mathbf{D} \in \mathbb{R}^{H \times W \times 1}}
\newcommand{\FeatureOutput}{\mathbf{F} \in \mathbb{R}^{\frac{H}{2^i} \times \frac{W}{2^i} \times C}}

\IEEEoverridecommandlockouts                              

\title{\LARGE \bf
Depth Matters: Exploring Deep Interactions of RGB-D for Semantic Segmentation in Traffic Scenes
}

\author{Siyu Chen$^{1}$*, Ting Han$^{2}$*, Changshe Zhang$^{3}$,Weiquan Liu$^{1}$, Jinhe Su$^{1}$$^{\dag}$, Zongyue Wang$^{1}$, Guorong Cai$^{1}$$^{\dag}$ \\
\textsuperscript{\rm 1} Jimei University, \textsuperscript{\rm 2} Sun Yat-sen University, 
\textsuperscript{\rm 3} Xidian University \\
\thanks{* Co-first author: Siyu Chen and Ting Han contribute equally.}
\thanks{$\dag$ Corresponding author: Jinhe Su, Guorong Cai. email: \{sujh, guorongcai\}@jmu.edu.cn).}%
}

\begin{document}

\maketitle 


\begin{strip}
    \centering
    \vspace{-25mm} 
    \parbox{\textwidth}{%
        \centering
        \includegraphics[width=1\textwidth]{fig/shift.pdf} 
        \captionof{figure}{Effectiveness of \textbf{Depth SAO} on attention and semantic segmentation quality.} 
        \label{fig:attention_shift} 
    }
\end{strip}


\begin{abstract}
RGB-D has gradually become a crucial data source for understanding complex scenes in assisted driving. However, existing studies have paid insufficient attention to the intrinsic spatial properties of depth maps. This oversight significantly impacts the attention representation, leading to prediction errors caused by attention shift issues. To this end, we propose a novel learnable Depth interaction Pyramid Transformer (\textbf{DiPFormer}) to explore the effectiveness of depth. Firstly, we introduce Depth Spatial-Aware Optimization (Depth SAO) as offset to represent real-world spatial relationships. Secondly, the similarity in the feature space of RGB-D is learned by Depth Linear Cross-Attention (Depth LCA) to clarify spatial differences at the pixel level. Finally, an MLP Decoder is utilized to effectively fuse multi-scale features for meeting real-time requirements. Comprehensive experiments demonstrate that the proposed DiPFormer significantly addresses the issue of attention misalignment in both road detection (+7.5\%) and semantic segmentation (+4.9\% / +1.5\%) tasks. DiPFormer achieves state-of-the-art performance on the KITTI (97.57\% F-score on KITTI road and 68.74\% mIoU on KITTI-360) and Cityscapes (83.4\% mIoU) datasets.

\end{abstract}

\section{INTRODUCTION}

With the rapid development of autonomous driving, the ability to accurately understand complex traffic scenes has become increasingly crucial. Traditional single-sensor data often fail short under varying lighting conditions and with dynamic objects. To overcome these challenges, combining texture information with spatial geometry, RGB-D based approaches \cite{piekenbrinck2024rgb,xia2024rgbd,Rajpal_2023_CVPR} have emerged as the mainstream methods. 

However, existing methods often treat depth information as auxiliary and overlook the intrinsic properties of depth maps\cite{chen2025hspformer}. As a result, conflicts between RGB and depth map in sensitive regions lead to misalignment in the learned attention maps, significantly affecting perception in complex real-world scenes. For instance, the attention shift shown in Fig.~\ref{fig:attention_shift}(a, b) leads to substantial prediction errors.


We believe that depth information plays a more crucial role in multi-modal semantic segmentation. To this end, we propose a novel learnable Depth interaction Pyramid Transformer (\textbf{DiPFormer}) to explore the effectiveness of depth. Firstly, we introduce Depth Spatial-Aware Optimization (\textbf{Depth SAO}) to represent spatial relationship characteristics. Secondly, we utilize Depth Linear Cross Attention (\textbf{Depth LCA}) to calculate the similarity of RGB-D. The interaction of feature spaces constructs the latent connections between images and the real world, and addresses the attention shift issues, as shown in Fig.~\ref{fig:attention_shift}(e). Multi-scale spatial relationships are generated to adapt to complex scenes across multiple scales. Finally, multi-scale features are further fused in the MLP decoder in a more efficient manner. 

Comprehensive experiments demonstrate that the DiPFormer significantly \textbf{provides the accurate attention} and improves the segmentation performance, as shown in Fig.~\ref{fig:attention_shift}(c, d). Specifically, Depth SAO significantly enhances segmentation performance, improving by 7.5\%, 4.9\%, and 1.5\% on datasets KITTI Road, KITTI-360, and Cityscapes, respectively (See Fig.~\ref{fig:attention_shift}(f)). Meanwhile, DiPFormer achieves state-of-the-art (SOTA) performance (\textbf{97.57\% F-score in KITTI Road, 68.74\% and 83.4\% mIoU in KITTI-360 and Cityscapes}, respectively) on the three datasets, further highlighting the importance of depth information. The contributions are as follows:


\begin{itemize}
    \item We propose a novel learnable Depth interaction Pyramid Transformer that sufficiently integrates spatial information to address attention drift issues.
    \item The interaction of feature spaces with spatial relationship characteristics constructs the latent connections between images and real world.
    \item The proposed method achieves Top-1 mIoU and F-score on the three well-known real world datasets with robustness and effectiveness.
\end{itemize}

\section{RALETED WORK}

\begin{figure*}
    \centering
    \includegraphics[width=\textwidth]{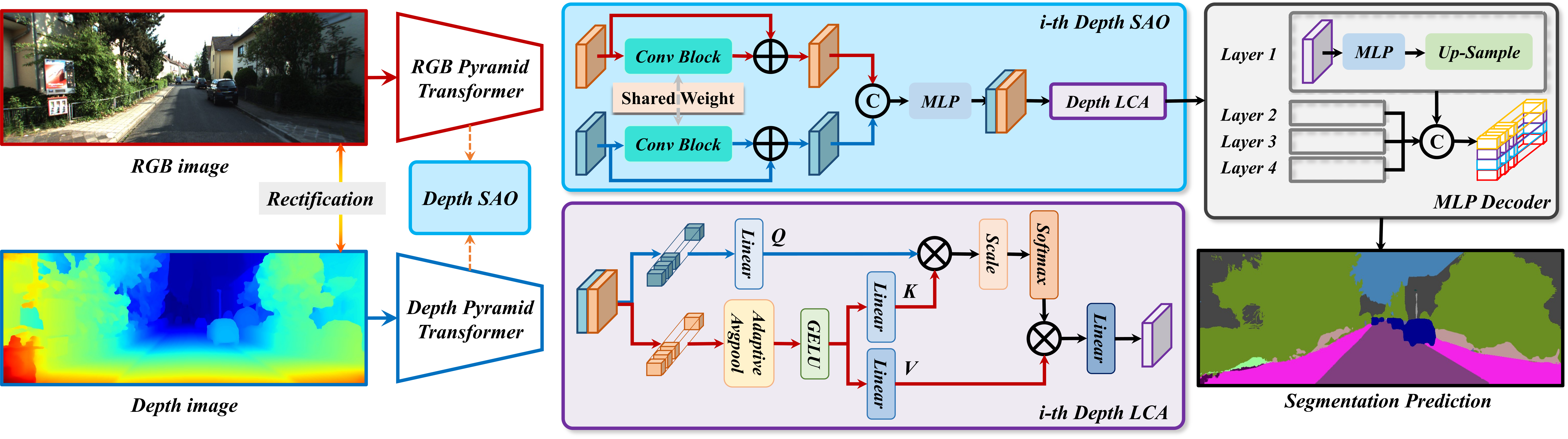}
    \caption{The overall framework of the proposed DiPFormer, which consists of layer-wise Depth SAO with Depth LCA, and MLP Decoder, for efficient and robust semantic segmentation.}
    \label{fig:model}
\end{figure*}

\textbf{Semantic Segmentation}. Previous semantic segmentation methods typically relied on the efficient CNN as backbone for feature extraction \cite{wang2023internimage,ji2023ddp,wu2023conditional}. However, due to the limitations of local representations, many methods adopt Vision Transformer for dense prediction \cite{erisen2024sernet,yang2024depth,jain2023oneformer}. To overcome the challenges that RGB images face under different weather and lighting conditions, multi-modal approaches have gained popularity and are now widely applied in the field of autonomous driving \cite{feng2020deep,ye2023fusionad,fu2024limsim++}. CMX \cite{zhang2023cmx} and CMNeXt \cite{zhang2023delivering} have introduced multiple attention modules to facilitate cross-modal learning between visual modalities. Nevertheless, these methods suffer from attention shift issues due to the neglect of the spatial properties of depth information. To this end, we propose Depth SAO and Depth LCA to correct attention shift without additional complex designs.

\textbf{Road Detection}. Road detection is a critical application of semantic segmentation in autonomous driving. Alongside the development of semantic segmentation, road detection also requires spatial information to assist with RGB feature extraction \cite{wang2021dynamic,chang2022fast, sun2019reverse}. RGB-L methods project point cloud onto the image plane for feature fusion \cite{gu2021cascaded,sun2022pseudo,khan2022lrdnet}, but they must address challenges related to computational complexity and registration issues. RGB-D methods extract and fuse features in parallel using CNNs or Transformer \cite{feng2024sne,han2024epurate,li2024roadformer}. These methods aim to provide richer feature representations but present errors where the two modalities conflict. Our goal is to explore the correlations between RGB-D features in the feature space to model the latent relationships between pixels and the real world.

\textbf{Position Embedding}. In Vision Transformer, position embeddings are essential for understanding the positions of image patches within a sequence. Traditional absolute position embeddings capture location information using fixed sine and cosine values \cite{vaswani2017attention,carion2020end,zhang2022dino,zhang2023vitaev2}, while relative position embeddings \cite{liu2022swin} represent the distance or difference between patches. Furthermore, many studies \cite{fan2024rmt,jiao2023dilateformer,fan2024lightweight,lu2024sbcformer,zhang2023vitaev2,zhang2024eatformer} have introduced convolutions to provide learnable implicit position embeddings. However, there is often a disconnect between these embeddings and the actual scene. To bridge this gap, we propose Depth SAO to establish an effective connection between position embeddings and the real world scenes.


\section{METHOD}

This paper introduces a novel learnable Depth interaction Pyramid Transformer (DiPFormer) for semantic segmentation. As shown in Fig.~\ref{fig:model}, DiPFormer consists of three main modules: Depth SAO, Depth LCA, and MLP decoder. DiPFormer formalizes the relationship between image feature spaces and spatial structure into depth spatial-aware. We details the designs as follows.

\subsection{DiPFormer}

Pyramid Transformer DiPFormer takes RGB images and depth maps of identical resolution as input. For an input image of size $H \times W \times 3$, the model first generates multi-level RGB and Depth features, with resolutions of $\frac{1}{2}$, $\frac{1}{4}$, $\frac{1}{8}$, and $\frac{1}{16}$, respectively. The extracted depth features are utilized as position embeddings in Depth SAO to address the attention shift caused by insufficient spatial guidance. This ensures effective alignment between the attention mechanism and the geometric structure of the scene. Subsequently, we employ Depth LCA to perform feature interaction in different feature spaces and obtain the latent spatial difference of pixels. Finally, the MLP decoder progressively restores these multi-level features with $H \times W \times N_{cls}$, where $N_{cls}$ represents the number of target classes. The predictions are constrained by the cross-entropy loss to acquire accurate semantic segmentation.

\begin{figure}[]
    \centering
    \includegraphics[width=0.5\textwidth]{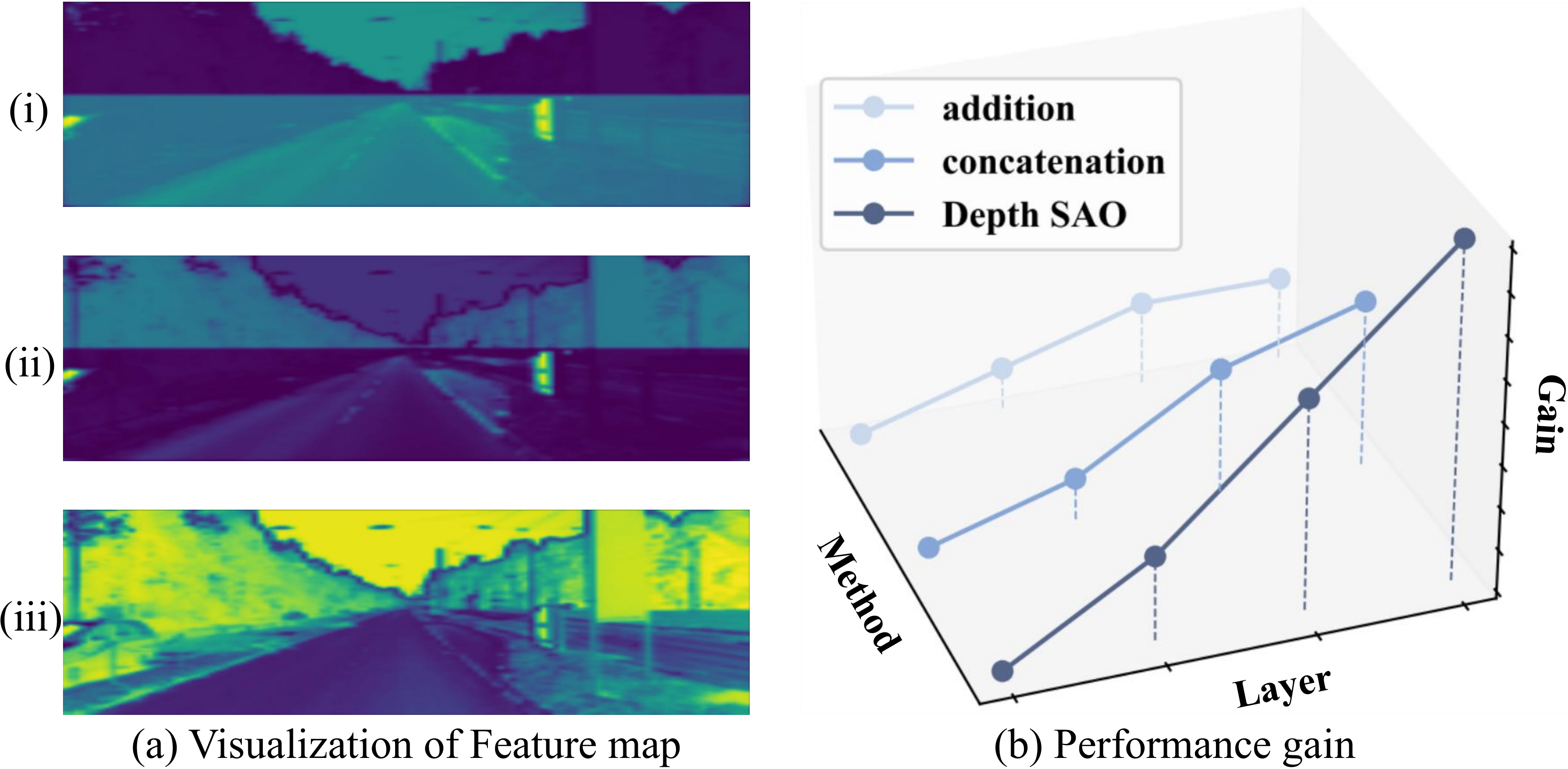} 
    \caption{Visualization the feature maps of different position embedding methods and the performance gain of these methods, where (i), (ii), and (iii) denote the pixel-wise addition, concatenation, and our proposed Depth SAO.} 
    \label{fig:feature}
\end{figure}

\begin{algorithm}[t!]
\fontsize{8pt}{10pt}\selectfont 
\captionsetup{font={small}}
\caption{Depth Spatial-Aware Optimization Computation}
\label{alg:depth_embedding}
\begin{algorithmic}[1]
\REQUIRE \parbox[t]{\dimexpr\linewidth-\algorithmicindent}{RGB input $\RGBinput$, \\ \textnormal{Depth input } $\Depthinput$}
\ENSURE \parbox[t]{\dimexpr\linewidth-\algorithmicindent}{Output feature $\FeatureOutput$ \textnormal{ for each layer}}

\FOR{each layer $l = 1, \dots, 4$}
    \STATE $\mathbf{R\_F}, \mathbf{D\_F} \gets \mathbf{R}, \mathbf{D}$
    \FOR{$i = 1$ to $3$}
        \STATE $\mathbf{R\_F}, \mathbf{D\_F} \gets \text{GN}(\text{Conv3x3}(\mathbf{R\_F}, \mathbf{D\_F}))$
        \IF{$i == 1$}
            \STATE $\mathbf{R\_F}, \mathbf{D\_F} \gets \text{MaxPooling}(\mathbf{R\_F}, \mathbf{D\_F})$
        \ENDIF
    \ENDFOR
    \STATE $\mathbf{F} \gets \text{Linear}(\mathbf{R\_F} + \mathbf{D\_F})$
\ENDFOR
\RETURN \ensuremath{\mathbf{F}}
\end{algorithmic}
\end{algorithm}

\subsection{Depth Spatial-Aware Optimization}

Unlike traditional position embeddings, we develop a more sophisticated depth position embedding based on depth features. We carefully consider potential issues such as sparsity and error accumulation that are able to arise during the depth map acquisition process. The position embeddings for pixel-wise addition and concatenation only consider biases, making it difficult to capture global positional relationships, as shown in Fig.~\ref{fig:feature}(a). Since element-wise addition and concatenation amplify errors through iterative attention, we introduce Depth SAO to effectively build spatial offset while ignore noise in the depth information. Error accumulation slows down performance gain, but the proposed DepthSAO overcomes this issue in Fig.~\ref{fig:feature}(b).

Specifically, we design a convolutional block (ConvBlock) with shared parameters to simultaneously learn RGB and depth features. On one hand, the ConvBlock performs Patch Embedding in the Vision Transformer to reduce feature resolution. On the other hand, ConvBlock learns the local RGB-D correspondences and extracts salient features through max-pooling. Subsequently, the RGB-D features are concatenated and fused through residual connections, and an MLP is used to model the spatial bias relationships. The process of depth SAO is described in Algorithm~\ref{alg:depth_embedding}. Depth SAO represents the spatial relationship characteristics between RGB-D in a learnable manner.

\subsection{Depth Linear Cross Attention}

We adopt cross-attention mechanism to more effectively manage the interaction between the two modalities. Depth LCA calculates the similarity of RGB-D feature spaces in non-local manner, thereby minimizing the risk of errors. Depth features as learnable position embedding provides valuable prior information about significant spatial relationships for feature extraction. Given the fused feature generated by Depth SAO, both RGB and depth features incorporate biases from the other modality. Therefore, we separate the fused features into new RGB and depth features for subsequent cross-attention.

In Depth LCA, we employ depth features as query and RGB features as key and value since spatial data provides distinctive features between pixels. Therefore, depth queries represent local spatial structural relationships while implicitly capturing the connection between pixels and the real world. Firstly, the depth feature $ D\_F $ is passed through a linear transformation to obtain the query matrix $ Q $. Concurrently, the RGB feature $ R\_F $ undergoes an average pooling operation, followed by a linear transformation to generate the key matrix $ K $ and the value matrix $ V $. Average pooling is used to extract general features from RGB images. This process is mathematically expressed as:
\begin{equation}
    \begin{aligned}
        Q &= Linear(D\_F); \\  
        K, V &= Linear(AvgPool(R\_F));
    \end{aligned}
\end{equation}

The attenntion effectively combines the global information from the RGB features with the local information from the depth features. In this formula, the dot product $ Q \cdot K^{T} $ between the query matrix $ Q $ and the key matrix $ K $ is computed to explore the similarity of RGB-D feature spaces. The scale factor $ scale $ is employed to balance the influence of features from different modalities. The similarity attention is then normalized using the $ softmax $ to generate the attention weight matrix. The weights are subsequently applied to the value matrix $ V $ to achieve the sufficient fusion of RGB-D feature spaces. It is denoted as:
\begin{equation}
    Attention(Q, K, V) = softmax(Q \cdot K^{T} \times scale) \cdot V
\end{equation}
Following the traditional Transformer architecture, we incorporate $num_{h}$ head spaces into Depth LCA. The attention from multiple heads is aggregated by $ \oplus $ to enhance feature representation.

Depth LCA effectively integrates RGB-D information and captures interactions from different feature spaces. The semantic features obtained through depth queries are represented as latent code with spatial and texture similarity. More importantly, the latent code creates a bridge between pixels and the real world, significantly mitigating attention shift issues. As a result, it alleviates the perception inconsistency in regions where modal information conflicts in traditional RGB-D fusion.

\subsection{MLP Decoder}

MLP Decoder is a lightweight decoder composed solely of MLP layers, avoiding excessive computationally overheads. Meanwhile, the MLP decoder has a larger and more efficient receptive field compared to convolutions. Depth SAO generates multi-scale highly local and non-local interaction features. Firstly, multi-layer features from the Depth SAO are compressed to a uniform number $ C $ of channels. Then, they are upsampled at the size of $ \frac{H}{4} \times \frac{W}{4} \times C $ individually. Thirdly, aggregation is performed by concatenation. Finally, an additional MLP layer is applied to predict the segmentation mask at the size of $ \frac{H}{4} \times \frac{W}{4} \times N_{cls} $ from the fused features. MLP decoder unifies the multi-scale features to produce complementary and powerful representations.

\subsection{Complexity}

In Depth LCA, adaptive average pooling reduces the feature dimension from $ H \times W \times C $ to $ P \times P \times C $. Consequently, the computational complexity of Depth LCA decreases from $ O(HWC) $ to $ O(P^{2}C) $, where $ P $ is much smaller than $H$ and $W$. Moreover, the computational complexity of MLP Decoder is $O(HWC_{in}M)$, which is lower compared to the traditional convolutional decoder's complexity of $O(HWC^{2})$, where $C_{in}$ and $M$ are much smaller than $C$. The reduced computational complexity ensures that our design is lightweight and efficient, effectively meeting the real-time demands of traffic scene understanding.



\section{EXPERIMENTS}

\subsection{Datasets and Metric}

\textbf{KITTI Road} \cite{wang2014color}: The KITTI Road Dataset includes 289 training RGB-D image pairs and 290 testing pairs, each with a resolution of $1,242 \times 375$ pixels. It covers three road scenarios: urban unmarked, urban marked, and urban multiple marked lanes. The Urban Road Benchmark provides the comprehensive road detection evaluation of the three aforementioned road scenes.

\textbf{KITTI-360} \cite{Liao2022PAMI}: KITTI-360 is a large-scale street scene dataset, and extends the original KITTI dataset \cite{Geiger2013IJRR}. It includes 49,004 training pairs and 12,276 testing pairs with a resolution of $ 1,408 \times 376 $. The dataset provides dense semantic segmentation annotations across 19 classes.

\textbf{Cityscapes} \cite{Cordts2016Cityscapes}: Cityscapes is an RGB-D dataset that is used for both road detection and semantic segmentation. It consists of 5,000 image pairs, divided into training, validation, and testing subsets (2,975 / 500 / 1,525). Each image is annotated across 19 categories with a resolution of $2,048 \times 1,024$.

\textbf{Metric}: The performance of road detection is primarily evaluated on the KITTI Road dataset using the Max F1-measure. In addition, metrics such as Precision (PRE), Recall (REC), Intersection Over Union (IoU), Average Precision (AP), False Positive Rate (FPR), False Negative Rate (FNR), and runtimes are also assessed. For the semantic segmentation tasks, the mean Intersection over Union (mIoU) was employed as the evaluation metric on the KITTI-360 and Cityscapes datasets.

\subsection{Implementation Details}

Our model training is conducted on four NVIDIA 4090 GPUs, with a batch size of 4 per GPU. Initially, the network model is pre-trained on ImageNet-1K \cite{deng2009imagenet}. During training, we apply various data augmentation techniques to enhance the model's generalization capability, including random flipping, random scaling within the range of $[0.5, 2]$, random color jittering, and random Gaussian blur. We opt for the AdamW optimizer with a weight decay of 0.05. The initial learning rate was set to 6e-5, and a cosine annealing strategy with a warm-up phase was employed for learning rate scheduling. 

\subsection{Comprehensive Evaluation}

\begin{table*}[t]
  \caption{Evaluation results in Urban Road Benchmark \cite{wang2014color}. The best results are in \textbf{bold}.}
  \label{tab: kitti road}
  \centering
  \resizebox{\textwidth}{!}{
  \begin{tabular}{c|c|cccccccc}
  \toprule
  Method & Modal & \textbf{\textcolor{red}{MaxF}}(\%) $\uparrow$ & AP(\%) $\uparrow$ & PRE(\%) $\uparrow$ & REC(\%) $\uparrow$ & FPR(\%) $\downarrow$ & FNR(\%) $\downarrow$ & Runtime(s) $\downarrow$ \\ 
  \midrule
  CLCFNet(2021)\cite{gu2021cascaded} & RGB-L& 96.38 & 90.85 & 96.38 & 96.39 & 1.99 & 3.61 &	\textbf{0.02} \\
  PLB-RD(2022)\cite{sun2022pseudo} & RGB-L& 97.42 & 94.09 & 97.30 & 97.54 & 1.49 & 2.46 & 0.46 \\
  3MT-RoadSeg(2023)\cite{milli2023multi} & RGB-L& 96.60 & 93.90 & 96.46 & 96.73 & 1.95 & 3.27 &	0.07 \\
  LRDNet+(2022)\cite{khan2022lrdnet} & RGB-L& 96.95 & 92.22 & 96.88 & 97.02 & 1.72 &2.98 & - \\
  \midrule
  SNE-RoadSeg(2020)\cite{fan2020sne} & RGB-D & 96.75 & \textbf{94.07} & 96.90 & 96.61 & 1.70 & 3.39 & 0.18 \\  
  NIM-RTFNet(2020)\cite{wang2020applying} & RGB-D& 96.02 & 94.01 & 96.43 & 95.62 & 1.95 & 4.38 & 0.05 \\ 
  DFM-RTFNet(2021)\cite{wang2021dynamic} & RGB-D& 94.78 & 94.05 & 96.62 & 96.93 & 1.87 & 3.07 & 0.08 \\  
  SNE-RoadSeg+(2021)\cite{wang2021sne} & RGB-D& 97.50 & 93.98 & 97.41 & 97.58 & 1.43 & 2.42 & 0.08 \\
  USNet(2022)\cite{chang2022fast} & RGB-D& 96.89 & 93.25 & 96.51 & 97.27  & 1.94 & 2.73 & \textbf{0.02} \\
  SNE-RoadSegV2(2023)\cite{feng2024sne} & RGB-D& 97.55 &	93.98 &	\textbf{97.57} & 97.53 & \textbf{1.34} & 2.47 & 0.03 \\
  EpurateNet(2024)\cite{han2024epurate} & RGB-D& 97.09 & 93.08 & 96.76 & 97.43 & 1.89 & 2.76 & \textbf{0.02} \\
  RoadFormer(2024)\cite{li2024roadformer} & RGB-D& 97.50 & 93.85 & 97.16 & 97.84 & 1.57 & 2.16 & 0.07 \\
  RoadFormer+(2024)\cite{huang2024roadformer+} & RGB-D& 97.56 & 93.74 & 97.43 & 97.69 & 1.42 & 2.31 & 0.04\\
  \midrule
  \rowcolor{gray!20}
  \textbf{DiPFormer (Ours)} & RGB-D& \textbf{\textcolor{red}{97.57}} & 92.94 & 97.34 & \textbf{97.79} & 1.47 & \textbf{2.21} & \textbf{0.02} \\  
  \bottomrule
  \end{tabular}
  }
\end{table*}

\begin{figure*}[t]
    \centering
    \includegraphics[width=\textwidth]{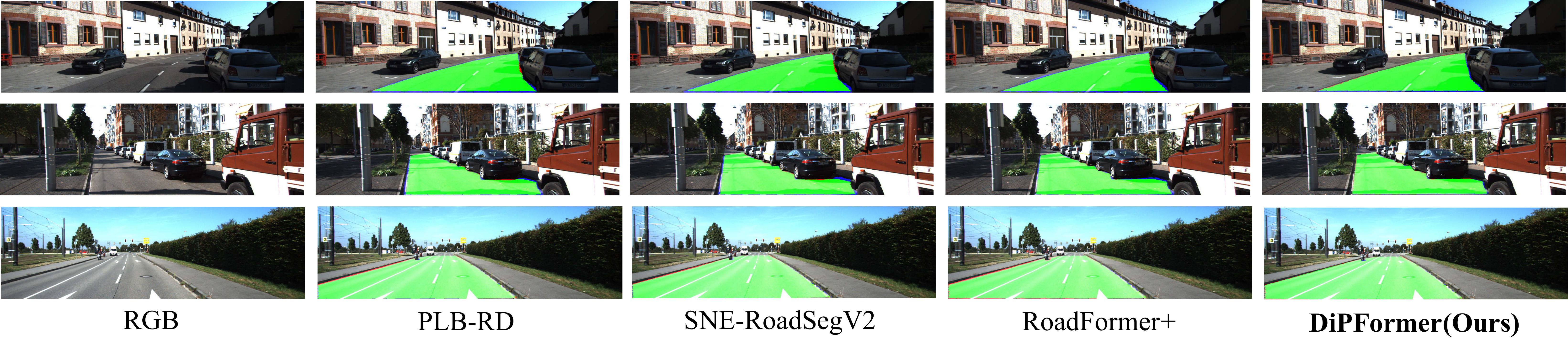} 
    \caption{Qualitative results on KITTI Road \cite{Liao2022PAMI} compared with latest high-performance methods.} 
    \label{fig:kitti road}
\end{figure*}

\begin{table}[t!]
	\centering
	\caption{Main results on Cityscapes \cite{Liao2022PAMI}, where mIoU is all categories evaluation and MaxF is road detection evaluation. The best results are in bold. } 
    \label{tab: Cityscapes results}
    \resizebox{0.5\textwidth}{!}{
	\begin{tabular}{c|c|c|c}
    \toprule
        Method & Modal & \textbf{\textcolor{red}{mIoU}}(\%) $\uparrow$ & MaxF (\%) $\uparrow$ \\
		\midrule
		    PVT(2021)\cite{wang2021pyramid} & RGB  & 78.6 & 93.74 \\ 
            SegFormer(2021)\cite{xie2021segformer} & RGB  & 81.0 & 94.67 \\
		    PVTv2(2022)\cite{wang2022pvt} & RGB & 80.6 & 95.51 \\
            HRViT(2022)\cite{gu2022multi} & RGB & 82.8 & - \\
            FAN-L-Hybrid(2023)\cite{zhao2023fully} & RGB  & 82.8 & - \\
            \midrule
            SA-Gate(2020)\cite{chen2020bi} & RGB-D  & 81.7 & - \\
            USNet(2022)\cite{chang2022fast} & RGB-D & - & 98.27 \\
            CMX(2023)\cite{zhang2023cmx} & RGB-D & 82.6 & 95.72 \\
            DFormer(2023)\cite{yin2023dformer} & RGB-D & 74.4 & - \\
            SNE-RoadSegV2(2024)\cite{feng2024sne} & RGB-D & - & 97.12 \\
            RoadFormer(2024)\cite{li2024roadformer} & RGB-D & 76.1 & 96.56 \\
            RoadFormer+(2024)\cite{huang2024roadformer+} & RGB-D & 77.4 & 97.96 \\
            \midrule
            \rowcolor{gray!20}
            \textbf{DiPFormer(Ours)} & RGB-D & \textbf{\textcolor{red}{83.4}} & \textbf{98.86} \\
		\bottomrule
	\end{tabular}
 }
\end{table}

\begin{table}[]
	\centering
	\caption{Main results on KITTI-360 \cite{Liao2022PAMI}. The best results are in bold. } \label{tab: kitti360 results}
    \resizebox{0.45\textwidth}{!}{
	\begin{tabular}{c|c|c|c}
    \toprule
        Method & Modal & Param(M) & \textbf{\textcolor{red}{mIoU}}(\%) $\uparrow$ \\
		\midrule
		  PVT(2021)\cite{wang2021pyramid} & RGB & 28.2 & 57.53  \\ 
            SegFormer(2021)\cite{xie2021segformer} & RGB & 25.9 & 61.37 \\
		  PVT v2(2022)\cite{wang2022pvt} & RGB & 29.1 & 59.70 \\
        \midrule
		  TokenFusion(2022)\cite{wang2022multimodal} & RGB-L & 26.0 & 54.55 \\
            CMX(2023)\cite{zhang2023cmx} & RGB-L & 66.7 & 64.31 \\
            CMNeXt(2023)\cite{zhang2023delivering} & RGB-L & 58.7 & 65.26 \\
        \midrule
            PGDENet(2022)\cite{zhou2022pgdenet} & RGB-D & 107.0 & 56.34 \\
            TokenFusion(2022)\cite{wang2022multimodal} & RGB-D & 26.0 & 57.44 \\
            CMX(2023)\cite{zhang2023cmx} & RGB-D & 66.7 & 64.43 \\
            CMNeXt(2023)\cite{zhang2023delivering} & RGB-D & 58.7 & 65.09 \\
            \midrule
            \rowcolor{gray!20}
            \textbf{DiPFormer(Ours)} & RGB-D & 56.3 & \textbf{\textcolor{red}{68.74}} \\
		\bottomrule
	\end{tabular}
 }
\end{table}

\begin{figure}[]
    \centering
    \includegraphics[width=0.35\textwidth]{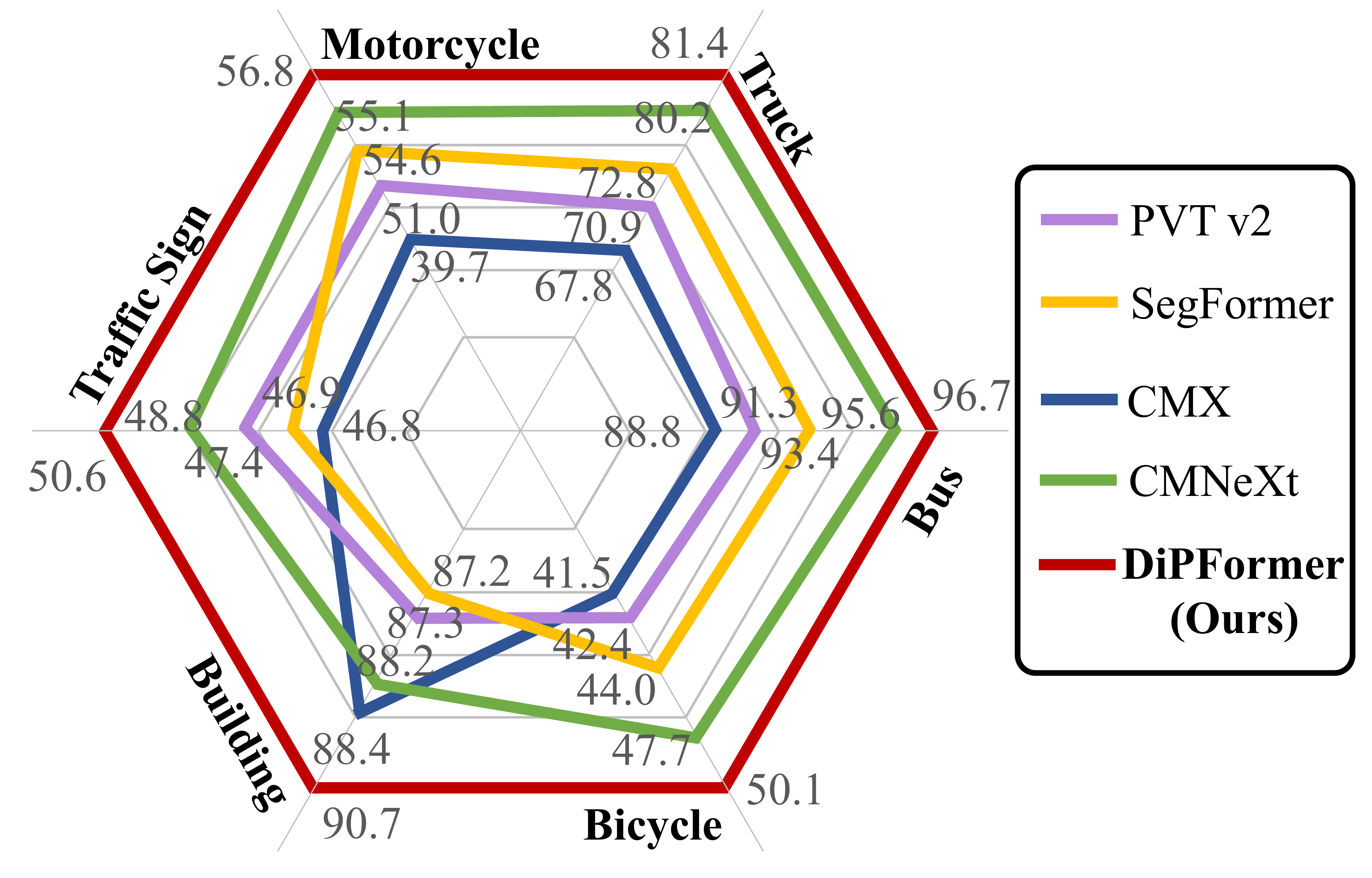} 
    \caption{Comparison of challenging categories in the KITTI-360 dataset \cite{Liao2022PAMI}.} 
    \label{fig:kitti360_iou}
\end{figure}

We conduct a comprehensive evaluation of the proposed DiPFormer using the benchmark datasets KITTI Road, KITTI-360, and Cityscapes, respectively. These datasets provide a diverse and challenging environment for assessing the performance in road detection and multi-class semantic segmentation tasks. Moreover, we compare DiPFormer against the current SOTA methods to thoroughly evaluate its advantages and areas of improvement. Note that RGB-L and RGB-D denote the input as RGB images with LiDAR point cloud and depth maps, respectively.

\subsubsection{\textbf{Road Detection}}

The Quantitative results on the KITTI Road dataset (see Tab.~\ref{tab: kitti road}) demonstrate that DiPFormer excels across multiple key metrics. Specifically, DiPFormer achieves the highest performance in terms of MaxF (97.57\%), REC (97.79\%), and FNR (2.21\%). Compared to previous method, DiPFormer maintains top performannce while also being highly competitive in terms of speed. Especially compared to Transformer-based methods, the speed is significantly improved by the efficient attention Depth LCA and MLP Decoder. Qualitative results are shown in Fig.~\ref{fig:kitti road}, where the scenes are UM, UMM, and UU from top to bottom. DiPFormer achieves the best edge segmentation in road and vehicles contours. Moreover, we conduct the road detection experiment on the Cityscapes dataset. As shown in Tab.~\ref{tab: Cityscapes results}, the proposed method achieves a best MaxF of 98.56\%, approximately 0.6\% higher than the second-best method.

\subsubsection{\textbf{Semantic Segmentation}}

We use Cityscapes and KITTI-360 dataset for the semantic segmentation evaluation. On the Cityscapes dataset (Tab.~\ref{tab: Cityscapes results}), DiPFormer delivers outstanding performance with 83.4\% mIoU, surpassing other advanced methods. The depth images from Cityscapes contain inherent noise, which inevitably affect the performance of RGB-D methods. However, the proposed method significantly mitigates error accumulation in depth images by leveraging the similarity in the RGB-D feature space and utilizing learnable depth position embeddings.

\begin{figure*}[t]
    \centering
    \includegraphics[width=\textwidth]{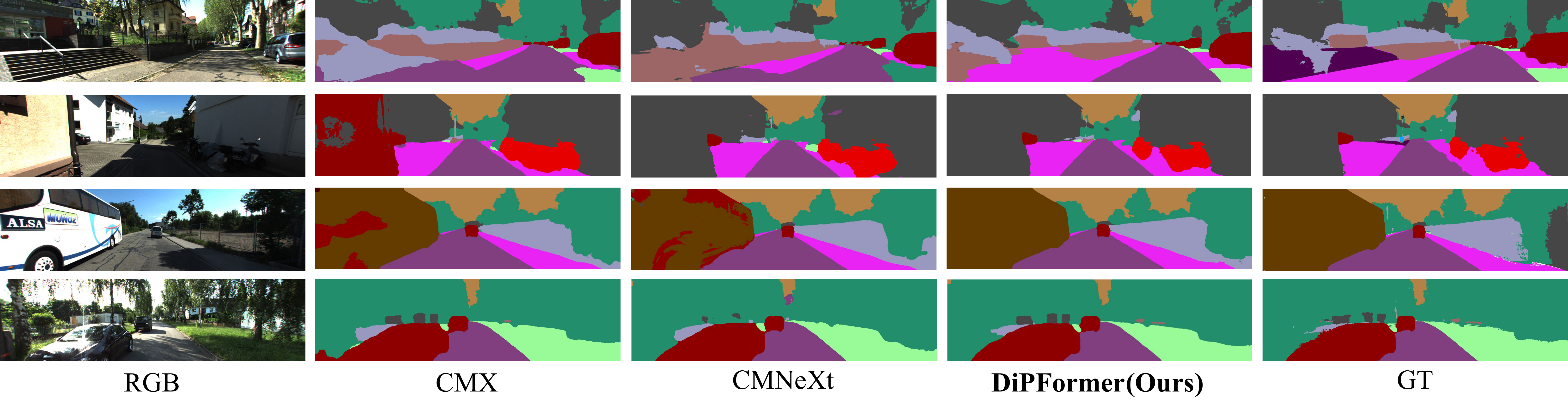}
    \caption{Qualitative results on KITTI-360 \cite{wang2014color} compared with latest high-performance methods.} 
    \label{fig:kitti360} 
\end{figure*}

In Tab.~\ref{tab: kitti360 results}, we assess the performance of DiPFormer on the KITTI-360 dataset. DiPFormer achieves a notable mIoU of 68.74\% with a modest parameter of 56.3M, significantly outperforming other methods such as CMNeXt (65.26\%) and CMX (64.43\%). Fig.~\ref{fig:kitti360_iou}, DiPFormer excels in several challenging categories, which are typically difficult to accurately identify in complex traffic scenarios. Fig.~\ref{fig:kitti360} demonstrates DiPFormer accurately segments the various objects in complex traffic environments.



\begin{table}[]
    \centering
    \caption{Ablation Study of DiPFormer Components.}
    \label{tab: Ablation}
    \begin{tabular}{c}
        \begin{subtable}{\linewidth}
            \centering
            \subcaption{Depth SAO \& LCA analysis.}
            \label{tab:subtable1}
            \resizebox{\linewidth}{!}{ 
            \begin{tabular}{c|ccc}
                \toprule
                Components & Param (M) & mIoU (\%) $\uparrow$ & mAcc (\%) $\uparrow$ \\
                \midrule
                baseline & 29.1 & 59.7 & 65.0 \\
                \midrule
                +Depth SAO & 47.8 & 63.1 & 70.2 \\
                +Depth LCA & 40.7 & 62.3 & 71.6 \\
                +\textbf{Depth SAO \& LCA} & 59.4 & \textbf{67.1} & \textbf{77.4} \\
                \bottomrule
            \end{tabular}
            }
        \end{subtable}
        \\
        \\
        \begin{subtable}{\linewidth}
            \centering
            \subcaption{MLP Decoder analysis.}
            \label{tab:subtable2}
            \resizebox{\linewidth}{!}{ 
            \begin{tabular}{c|ccc}
                \toprule
                Components & Param (M) & mIoU (\%) $\uparrow$ & mAcc (\%) $\uparrow$ \\
                \midrule
                baseline & 29.1 & 59.7 & 65.0 \\
                \midrule
                w/ MLP Decoder & 25.3 & 61.4 & 69.0 \\
                DiPFormer w/o MLP Decoder & 59.4 & 67.1 & 77.4 \\
                \textbf{DiPFormer w/ MLP Decoder} & 56.3 & \textbf{68.7} & \textbf{79.8} \\
                \bottomrule
            \end{tabular}
            }
        \end{subtable}
    \end{tabular}

\end{table}

\begin{table}[]
    \vspace{-\baselineskip}
	\centering
	\caption{Semantic segmentation performance of different position embeddings on KITTI-360 val set. The best results are in bold. We apply different PEs to the PVT v2-B2 \cite{wang2022pvt} to ensure fairness in the comparison.
    \label{tab: DepthEmbed}}
    \resizebox{0.45\textwidth}{!}{
	\begin{tabular}{c|cc}
		\toprule
    Position Embeddings & Input & mIoU(\%) $\uparrow$ \\		
		\midrule
		Sine / Cosine PE\cite{dosovitskiy2020image,vaswani2017attention} & RGB & 57.4  \\
		Learnable PE\cite{wang2021pyramid,yuan2021tokens} & RGB-D & 59.1 \\
		Implicit PE\cite{wang2022pvt,xie2021segformer} & RGB-D & 62.3 \\ 
    \midrule
        \rowcolor{gray!20}
		\textbf{Depth SAO (Ours)}  & RGB-D & \textbf{66.3} \\
		\bottomrule
	\end{tabular}
 }
 \vspace{-2.0em}
\end{table}

\subsection{Ablation Studies}

\textbf{Components analysis of DiPFormer.} To evaluate the individual contributes of our designs, we incrementally added each component and assessed their impact. As shown in Tab.~\ref{tab: Ablation}, the Depth SAO and Depth LCA contribute significantly to the performance improvement. Especially when built upon the SAO, LCA further enhances the interaction within the RGB-D feature space and provide mIoU increasing from 63.1\% to 67.1\%. The MLP Decoder reduces the parameter by approximately 3\% and improves mIoU by 2\% compared with Vanilla. These results demonstrate the efficiency of the MLP Decoder and its ability to provide richer feature representations.

\textbf{The effectiveness of Depth SAO.} Tab.~\ref{tab: DepthEmbed} compares performance across different position embeddings (PE) on the KITTI-360 validation set. The traditional Sine / Cosine PEs achieve an mIoU of 57.4\%, the learnable PE reaches 59.1\%, and the implicit PE improves to 62.3\%. The proposed Depth SAO significantly increases the mIoU to 66.3\%, outperforming all other methods ad demonstrating the superiority in capturing spatial relationship.

\section{CONCLUSIONS}

This paper presents the DiPFormer framework, a novel approach for semantic segmentation and road detection in complex real-world traffic environment. First, we develop Depth SAO as learnable position embedding to represent the spatial relationship. Next, we design the Depth LCA to generate the similarity of RGB-D feature spaces and clarify the spatial difference at the pixel level. Lastly, the multi-scale features are efficient fused to provide the dense predictions. Extensive experiments have verified the effectiveness of our designs. Notably, our method constructs the implicit spatial representation to model the correlation between pixels and real world scenes. Our method surpasses previous approaches both on semantic segmentation and road detection tasks in three well-known datasets, demonstrating the advancement of our approach.

\textbf{Limitations.} One notable limitation of our work is still relies on the quality of the depth images to some extent. Despite our design mitigates error accumulation, the improvement contributed by sparse depth is not significant, as shown in Fig.~\ref{fig:attention_shift}(f: Cityscapes). Therefore, we will explore an unified framework that integrates segmentation with depth estimation to reduce reliance on depth images and strengthen the relationship between pixels and depth.

\section*{ACKNOWLEDGMENT}
This work was supported in part by the Natural Science Foundation of Xiamen, China, under Grant 3502Z202373036; in part by the National Natural Science Foundation of China under Grant 42371457, Grant 42301468, and Grant 42371343; in part by the Key Project of Natural Science Foundation of Fujian Province, China, under Grant 2022J02045; in part by the Natural Science Foundation of Fujian Province, China, under Grant 2022J01337, Grant 2022J01819, Grant 2023J01801, Grant 2023J01799, Grant 2022J05157, and Grant 2022J011394.







\bibliographystyle{IEEEtran}
\bibliography{IEEEabrv}

\end{document}